\documentclass[11pt]{article}
\usepackage[T1]{fontenc}
\usepackage{amsmath}
\usepackage{amssymb}
\usepackage{graphicx}
\usepackage{booktabs}
\usepackage{float}
\usepackage{tabularx}
\usepackage{hyperref}
\usepackage{xcolor}
\usepackage{enumerate}
\usepackage{multirow}

\usepackage[margin=1in]{geometry}

\makeatletter
\renewcommand{\@maketitle}{%
  \begin{center}%
    \vspace*{-1cm}
    \rule{\linewidth}{2pt} \\[0.5cm]
    {\Large \scshape \@title} \\[0.5cm]
    \rule{\linewidth}{2pt} \\[1.5cm]
    {\@author}
  \end{center}
  \vspace{1cm}
}
\makeatother

\title{Open-Set Cattle Muzzle Identification:\\
A Leakage-Controlled Benchmark and Evaluation Protocol}

\author{
\begin{tabular*}{\textwidth}{@{\extracolsep{\fill}}ccc@{}}
\textbf{Lalit BC} & \textbf{Dharmendra Singh Chaudhary} & \textbf{Shovit Nepal} \\
Map Mentors & Department of Electronics & School of Environmental, \\
Lalitpur, Nepal & and Computer Engineering & Civil, Agricultural and \\
Fort Valley State University & Tribhuvan University, Nepal & Mechanical Engineering \\
Fort Valley, GA, USA & & University of Georgia, USA \\
lbc@wildcat.fvsu.edu & dharmendra.078bct016@tcioe.edu.np$^*$ & shovit.nepal@uga.edu \\[0.5cm]
\multicolumn{3}{c}{$^*$Correspondence: dharmendra.078bct016@tcioe.edu.np}
\end{tabular*}
}

\date{\today}

\begin{document}

\maketitle

\begin{abstract}
Reliable individual cattle identification supports disease surveillance, vaccination records, and livestock insurance. Although the bovine muzzle offers a stable, non-contact biometric, existing systems assume a closed set of enrolled animals, limiting practical deployments. We therefore reformulate cattle muzzle biometrics as an open-set, gallery-based identification problem that rejects unseen animals and enables incremental enrollment without retraining. We introduce a leakage-controlled evaluation protocol using identity-disjoint splits, per-fold retraining, held-out threshold calibration, verified duplicate removal, and bootstrap confidence intervals. We evaluate the framework with two contrasting embedding configurations: a hybrid CNN--ViT metric-learning model and the MegaDescriptor-L foundation model. Under oracle thresholds, the hybrid model achieves detection-and-identification rates of 98.3\%, 96.4\%, and 93.6\% at false-accepted targets of ($10^{-1}$, $10^{-2}$, $10^{-3}$) respectively, while MegaDescriptor-L achieves 99.3\%, 98.1\%, and 96.1\%. However, deployable calibration reveals an important gap: the hybrid model achieves 1.03\% FAR at a 1\% target, whereas MegaDescriptor-L reaches 2.44\%. Incremental enrollment further achieves Rank-1 accuracy above 91\% with a single image and up to 97.3\% with eight images, without retraining or degrading the existing gallery. These findings highlight threshold calibration and embedding quality as key factors for practical open-set cattle identification.
\end{abstract}


\section{Introduction}

Individual animal recognition is critical to livestock management. Disease surveillance, vaccination and treatment records, breeding decisions, and insurance claims all rest on reliable individual identification, and identification error may propagate into every record that follows. The global cattle population is approximately 1.57 billion head (FAOSTAT)~\cite{fao2025}, and the United States cattle and calf sector reported cash receipts of \$101.1 billion in 2023 (USDA ERS)~\cite{usda_ers}, so even small gains in identification reliability matter at scale.

Most farms still identify animals using physical methods such as ear tags, tattoos, brands, and RFID transponders. These are simple to apply but fragile in practice~\cite{awad2016,wardrope1995}. Quality of ear tags, scratches from farm facilities, and ill intention of the people lead ear tags to be lost, torn, or swapped easily~\cite{wardrope1995, Wang2024}; tag transfer between animals has also been reported as a specific cause for insurance fraud and moral hazards in buffalo and cattle schemes~\cite{Kim2026}. Hot iron branding is painful to animals, and it also harms the animal-human relationship~\cite{DeOliveira2024}, and RFID needs reader hardware at fixed points and is costly for smallholders~\cite{ruhil2013,islam2024}. Despite these challenges, RFID and ear tags are widely used identification methods in modern livestock farming and are useful if implemented carefully. On the other hand, cattle muzzle avoids these problems. Its pattern of beads and ridges has been reported as individually distinctive and relatively stable, much like a human fingerprint~\cite{barry2007,petersen1922}, and it can be photographed with a phone. Deep learning has pushed muzzle recognition to very high accuracy~\cite{li2022,kaur2022,shojaeipour2021,hossain2022}, in close-set and dataset-specific experimentation. Some of the recent work even fuses convolutional and transformer features to improve it further~\cite{dulal2025}.

Nearly all of this work makes one assumption that is rarely stated: it is closed-set. Every query animal is assumed to be already enrolled, so the model only has to pick the best-matching known identity. Real herds do not work this way. New purchases, transferred animals, animals missing, and unenrolled calves appear constantly, and a closed-set model cannot say ``I have not seen this one''; it simply returns the nearest match with confidence~\cite{Salehi2021, Zhou2021}, corrupting the record. The task that actually matters is an open-set approach that identifies the animals that are enrolled, rejects the ones that are not, and adds new animals to the gallery without retraining.

This is a different learning objective. A closed-set classifier only needs to predict a known label, whereas open-set identification needs an embedding space with a usable geometry, known identities in tight clusters, unknown animals outside them, and room for a rejection threshold to sit. That is why we take a metric-learning view rather than a classification view.

Some of the studies have brought open-set recognition to livestock, but not to the muzzle on its own. Meng et al.~\cite{meng2023} handle known--unknown recognition on cattle faces; Wang et al.~\cite{wang2024} work from overhead and coat images; and Kumar et al.~\cite{kumar2025}, the closest to us, fuse face and muzzle for one-to-one verification and argue that the muzzle alone is not discriminative enough. Pig-face variants also exist~\cite{wang2023,ma2025}. Two things are missing across this work: none evaluates the muzzle by itself in an open-set identification setting, and none reports detection at strict false-accept rates under a protocol that controls for identity leakage. That control is not a detail. Open-set metrics are extremely sensitive to it. An early version of our own pipeline reported a detection rate at FAR=$10^{-3}$ of $23.8\pm20.9\%$, caused entirely by protocol defects rather than the model. Leakage of this kind, evaluating ``unknown'' animals that were already seen during model development, is exactly what a careful protocol must prevent, and is a documented risk in the closely related open-set literature more broadly~\cite{scheirer2013}.

This paper makes three contributions:
\begin{enumerate}
\item We reformulate cattle muzzle identification as an \textbf{open-set, gallery-based problem} that rejects unseen animals and enrolls new ones without retraining.
\item We introduce a \textbf{leakage-controlled evaluation protocol}, identity-disjoint splits, per-fold retraining, held-out threshold calibration, verified duplicate removal, and bootstrap confidence intervals, and release it as, to our knowledge, the first publicly released leakage-controlled benchmark protocol for this task. This is the primary contribution.
\item We \textbf{validate the framework across two contrasting embedding configurations}, a hybrid CNN--ViT metric-learning model built for this work and the MegaDescriptor-L foundation model, and show the formulation and protocol hold across both.
\end{enumerate}

Under this protocol, both embeddings clear 90\% detection at every operating point (Table~\ref{tab:headline}). Our claim is deliberately narrow. We do not claim the first open-set method for cattle, nor the first CNN--Transformer hybrid for the muzzle~\cite{dulal2025}; rather, we claim the first publicly released, open-set, gallery-based benchmark protocol for muzzle identification, and the finding that the muzzle alone supports reliable rejection at strict false-accept rates. We also find that a plain cosine score is enough, given a well-trained embedding.

\section{Related Work}\label{sec:related}

\subsection{Muzzle Biometrics and CNN--Transformer Hybrids}

The muzzle has been treated as a fingerprint-like biometric for a century~\cite{barry2007,petersen1922}. Early systems paired handcrafted descriptors (SIFT, SURF, ORB) with classical classifiers and reached around 90\% accuracy on small sets~\cite{kaur2022}. Deep learning improved this sharply; Li et al.~\cite{li2022} released the Beef Cattle Muzzle Database and benchmarked convolutional networks to 98.7\% accuracy, with similar results on mixed breeds~\cite{shojaeipour2021} and Hanwoo cattle~\cite{lee2023}. Most recently, Dulal et al.~\cite{dulal2025} fuse CNN and transformer features with multi-head attention (MHAFF) for muzzle identification, exceeding 99\%. All of these systems are closed-set. They report classification accuracy among identities seen in training and do not measure the ability to reject an unenrolled animal. Our work targets that missing capability rather than a better fusion module.

\subsection{Open-Set Recognition}

Open-set recognition asks a classifier to reject inputs from unseen classes. Scheirer et al.~\cite{scheirer2013} formalized open-set risk; Bendale and Boult~\cite{bendale2016} introduced OpenMax. Later methods add losses that reserve mass for the unknown~\cite{dhamija2018}, surveyed in~\cite{geng2020}. Vaze et al.~\cite{vaze2022} showed that a well-trained embedding is itself a strong open-set detector, which we confirm for the muzzle. Performance is measured with the detection-and-identification rate at fixed false-accept rates (DIR@FAR), the open-set classification-rate curve (OSCR), and the equal-error rate (EER). These measures are standard in face recognition, speaker verification, and person re-identification, from which we borrow angular-margin embeddings, cohort score normalization, and threshold-based rejection, but they have not been used for the muzzle.

\subsection{Open-Set Recognition in Animal Biometrics}

Because herds gain and lose animals on a farm, open-set recognition is a natural fit for livestock, and a few studies have adopted it. Andrew et al.~\cite{andrew2021} reject Holstein-Friesian cattle by coat pattern; Meng et al.~\cite{meng2023} use reciprocal-point learning on cattle faces; Wang et al.~\cite{wang2024} use overhead views; and pig-face systems range from attention embeddings~\cite{wang2023} to a vision transformer that registers new pigs without retraining~\cite{ma2025}. Kumar et al.~\cite{kumar2025} fuse face and muzzle for verification. In all cases, the reported operating point is a single OSCR or AUROC value on a small unknown set, without DIR@FAR at controlled false-accept rates and without a demonstrated leakage-controlled protocol. No prior study combines the muzzle modality, open-set identification, strict-FAR reporting, and leakage control.

\subsection{Metric Learning}

Metric learning maps images to a space where same-identity images are close and different identities far apart, so enrollment is just storing a vector and matching is a distance comparison. Angular-margin losses shape this space, and ArcFace~\cite{deng2019} adds a fixed angular margin, AdaFace~\cite{kim2022} scales the margin by image quality via the feature norm, and supervised contrastive learning~\cite{khosla2020} optimizes neighborhood structure directly. Our hybrid embedding combines a CNN~\cite{he2016} and a ViT~\cite{dosovitskiy2021} with GeM pooling~\cite{radenovic2018} and a BNNeck head~\cite{luo2019}. The second embedding, MegaDescriptor~\cite{cermak2024}, is a Swin-based~\cite{liu2021} foundation model for animal re-identification.

\section{Materials and Methods}

\subsection{Overview}

A network turns each muzzle image into a 512-dimensional embedding on the unit sphere, trained so that images of the same animal sit close together. An animal is enrolled by averaging a few of its embeddings into a prototype and storing it in a gallery; no retraining is needed. A new image (a probe) is compared to every prototype by cosine similarity. If the best match is above a threshold, the system returns that identity; otherwise, it reports the animal as unknown. That reject step is the whole point, and it is what closed-set systems lack. Enrolling a new animal later is the same operation as before, storing one more prototype.

\begin{figure}[H]
\centering
\includegraphics[width=0.9\textwidth]{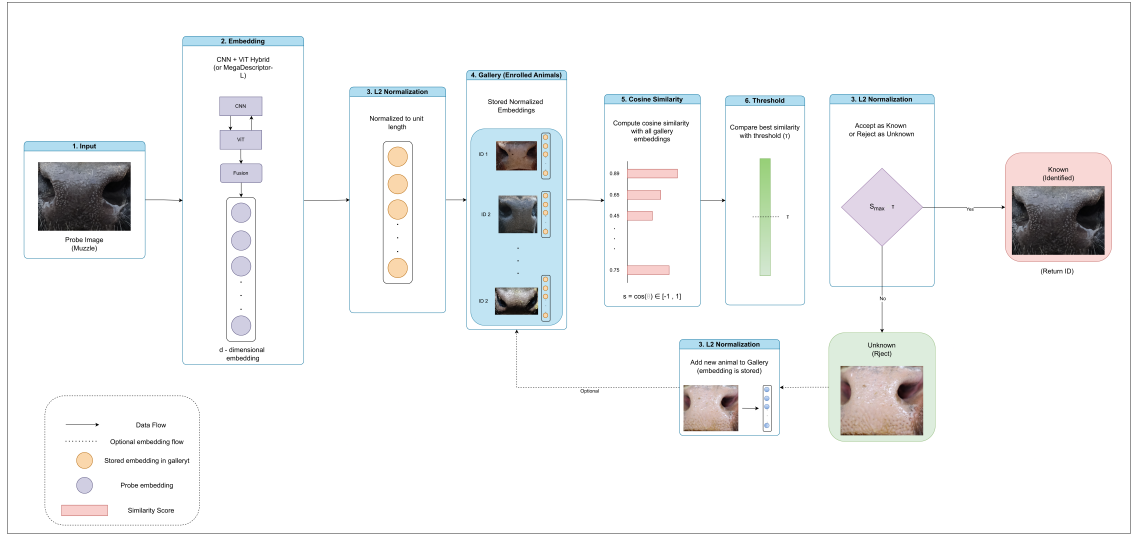}
\caption{System pipeline overview: (top) muzzle image acquisition and preprocessing, (middle) hybrid CNN-ViT embedding model, (bottom) open-set identification with enrollment and threshold-based rejection.}
\label{fig:system_pipeline}
\end{figure}

\subsection{Dataset}

We use the public Beef Cattle Muzzle/Nose-print Database~\cite{li2022,xiong2022}: 268 US feedlot cattle (Angus, Angus $\times$ Hereford, and Continental $\times$ British cross), of 4923 images, captured in 2021 with a 26 MP camera and already cropped to the muzzle, 18.4 images per animal on average (median 16, range 4--70). Two cleaning steps are applied to the dataset, in order: identities with fewer than four images are dropped so every retained animal can supply disjoint training, gallery, and probe images, and verified duplicate identities (280 images; the same physical animal filed under two folder names) are removed, leaving 249 identities and 4643 images. Duplicates were flagged by camera-frame provenance and confirmed by consensus visual review; the full evidence is released as \texttt{muzzle\_duplicates\_final.html}. This is a single herd at one site, so it supports a rigorous internal benchmark but not a claim of cross-farm generalization.

\begin{table}[h]
\centering
\caption{Dataset provenance.\label{tab:dataset}}
\begin{tabularx}{0.6\textwidth}{Xcc}
\toprule
\textbf{Stage} & \textbf{Identities} & \textbf{Images}\\
\midrule
Original database~\cite{xiong2022} & 268 & 4923\\
$-$ fewer than 4 images per identity & 0 & 0\\
$-$ verified duplicate identities & 19 & 280\\
\textbf{Retained benchmark} & \textbf{249} & \textbf{4643}\\
\bottomrule
\end{tabularx}
\end{table}

\subsection{Open-Set Formulation}

Each image maps to an $\ell_2$-normalized embedding $f_\theta(x)\in\mathbb{R}^{512}$. Identity $j$ is stored as a prototype $g_j$, the normalized mean of its gallery embeddings:
\begin{equation}
g_j=\bar e_j/\lVert \bar e_j\rVert_2,\qquad \bar e_j=\tfrac{1}{|\mathcal{G}_j|}\sum_{x\in\mathcal{G}_j} f_\theta(x).
\end{equation}

A probe $p$ scores $s(p)=\max_j \langle f_\theta(p),g_j\rangle$. If $s(p)\ge\tau$ the system returns $\arg\max_j\langle f_\theta(p),g_j\rangle$; otherwise $p$ is rejected as unknown. The embedding must therefore both rank the correct identity first and keep known and unknown probes on opposite sides of $\tau$.

\subsection{Hybrid CNN--ViT Embedding}

We build the embedding from two branches. A ResNet-50~\cite{he2016} captures local texture, the fine ridges. Its final residual stage produces a $12\times12\times2048$ feature map from a $384\times384$ image, flattened row-major into 144 region tokens. A ViT-Small/16~\cite{dosovitskiy2021}, run at $384\times384$ with bi-cubically-interpolated positional embeddings, captures global shape as 576 patch tokens plus one class token (577 total, width 384). The class token is retained and fused as an ordinary token. Both branches are projected to width 512 and fused via bidirectional cross-attention over two layers, 8 heads (head dimension 64), pre-LayerNorm, residual connections, feed-forward width 1024 (2$\times$ expansion, GELU), dropout 0.1. A quality gate then decides, per image and per dimension, how much to trust each branch. The gated vector is added to a GeM-pooled residual and passed through a BNNeck head to the final 512-D embedding. The pre-normalization vector norm is kept as a quality signal, which scales the AdaFace margin during training, and is smaller for unfamiliar animals, though after $\ell_2$ normalization it plays no role in the plain-cosine headline score.

These components (cross-attention fusion, gated fusion, norm-based quality weighting) are all established, and CNN--Transformer fusion has already been used for the muzzle~\cite{dulal2025,kumar2025}. It is designed to provide the open-set formulation with a strong embedding, not to advance fusion design.

\begin{figure}[H]
\centering
\includegraphics[width=0.85\textwidth]{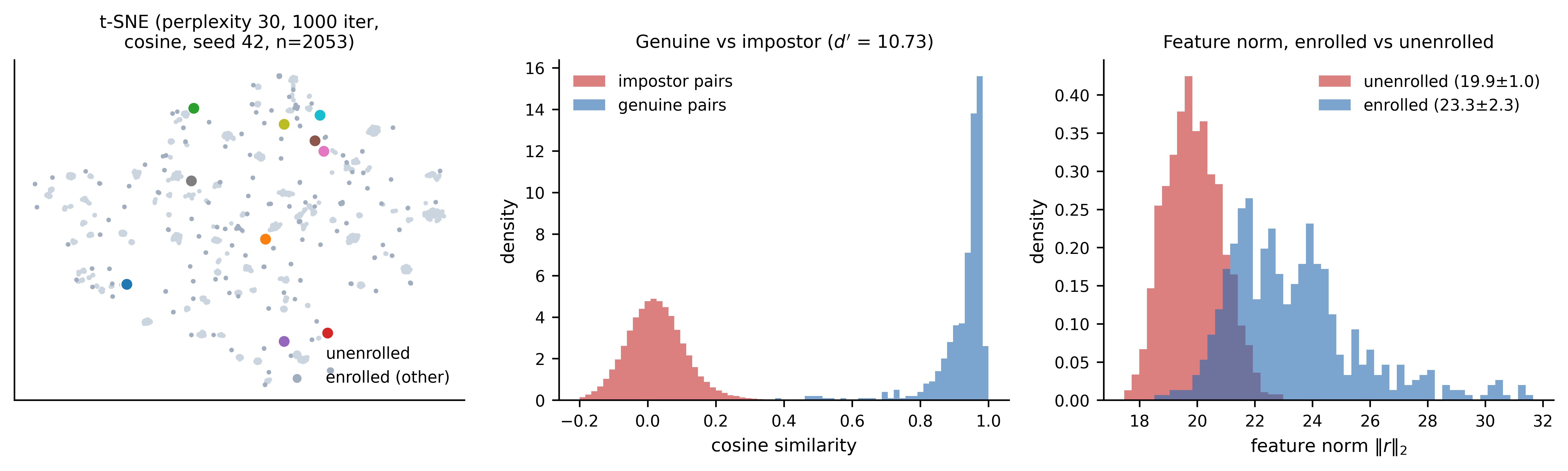}
\caption{Hybrid CNN-ViT embedding architecture: (left) dual-branch feature extraction with cross-attention fusion, (right) gated pooling and quality weighting for robust embedding.}
\label{fig:embedding_architecture}
\end{figure}

\subsection{Data Augmentation}

Every image is first passed through CLAHE on the L channel of CIELAB (clip limit $2.0$, tile grid $8\times8$) to normalize exposure without altering hue. During training, the normalized image undergoes geometric transforms (random resized crop, horizontal flip, rotation, affine shear, perspective distortion), photometric jitter (brightness, contrast, saturation, hue, sharpening, grayscale), and degradation noise (Gaussian blur, motion blur, JPEG compression). Two occlusion strategies add robustness: random erasing ($p=0.20$, area $[0.02,0.25]$) and CutMix~\cite{yun2019} ($p=0.20$, $\alpha=1.0$). At evaluation time, only the deterministic CLAHE--resize--normalize pipeline is applied. For inference, we additionally average embeddings over a 5-view test-time augmentation ensemble (direct, horizontal flip, center crop, two corner crops at $1.1\times$ up-scaling).

\subsection{Training and Scoring}\label{sec:training}

The embedding is trained with a sub-center AdaFace loss~\cite{kim2022} (scale 64, margin 0.5), which scales its angular margin by the feature norm so that sharp images are pushed harder than blurred ones; the same norm becomes the rejection cue at test time. Two auxiliary terms help: a center loss~\cite{wen2016} tightens clusters and a supervised-contrastive term~\cite{khosla2020} shapes the cosine neighborhood. We select checkpoints on val-unknown using a composite metric. For the second embedding, we simply replace the backbone with MegaDescriptor-L~\cite{cermak2024} (trained with ArcFace~\cite{deng2019}) and keep everything else identical, so that any difference reflects the backbone-plus-loss configuration as a whole, not the backbone alone.

At test time, there is no classifier. We score by cosine similarity to prototypes and threshold. Writing $s_1\ge s_2\ge\dots$ for the sorted prototype similarities of a probe, plain cosine is $s_1$; margin is $s_1+\lambda(s_1-s_2)$, $\lambda=0.5$; entropy is $s_1\cdot\exp(-\alpha H)$; combined multiplies the entropy gate by the margin score; AS-norm~\cite{matejka2017} centers $s_1$ by the probe's and gallery's top-$K$ cohort similarity means ($K=40$, best match excluded), without dividing by cohort $\sigma$; OpenMax~\cite{bendale2016} fits per-class Weibull tails (tail size 3, $\alpha=5$).

\subsection{Leakage-Controlled Protocol}\label{sec:protocol}

Open-set metrics are highly sensitive to identity leakage and test-set tuning; a naive version of our protocol produced a FAR=$10^{-3}$ detection rate of $23.8\pm20.9\%$, driven by three compounding defects: (i) identities were re-partitioned after a single model had already been trained on one partition, so ``unknown'' probes sometimes included identities the model had trained on; (ii) checkpoints were selected using DIR@FAR on the same probes used for the final report; and (iii) at strict FAR the threshold is fixed by only 1--2 unenrolled scores, so a single duplicate identity can pin it. The protocol below removes all three: identities are split once per fold into three identity-disjoint pools, each fold trains its own model, and duplicates are removed before splitting. Known ($\approx$50\%, 124 identities): training and gallery enrollment. Val-unknown ($\approx$20\%, 50 identities): checkpoint selection and threshold calibration only, never enrolled or trained on. Test-unknown ($\approx$30\%, 75 identities): touched exactly once per fold, for the final report only. Within each known identity, images are split 50\% training / 25\% gallery / 25\% probe.

\begin{figure}[H]
\centering
\includegraphics[width=0.8\textwidth]{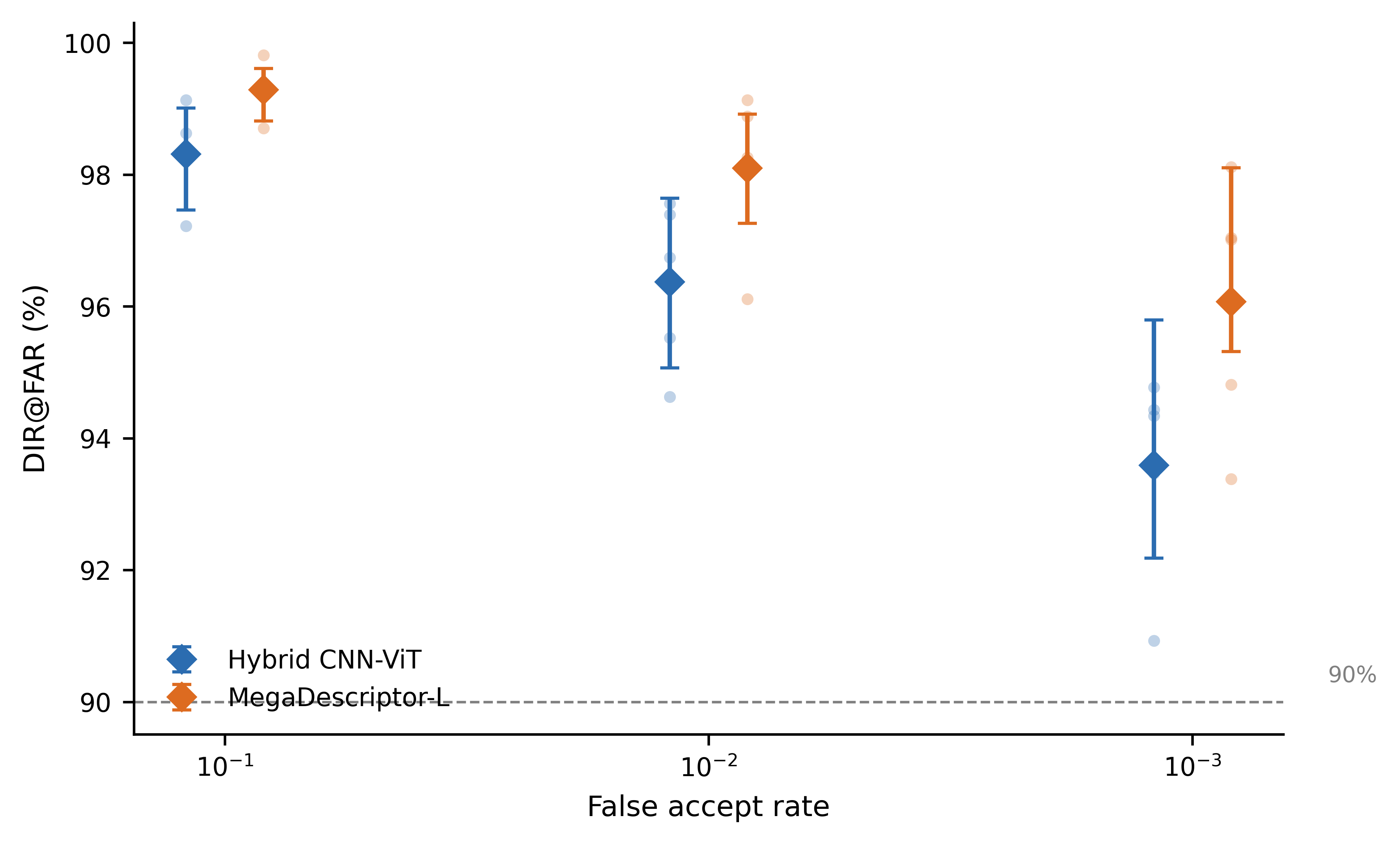}
\caption{Leakage-controlled protocol with identity-disjoint splits: known (124), validation-unknown (50), and test-unknown (75) identities are partitioned separately to prevent information leakage and ensure valid evaluation at strict false-accept rates.}
\label{fig:protocol_partition}
\end{figure}

The protocol runs over five repeated identity-disjoint partitions (a fresh model per partition). DIR@FAR at $\{10^{-1},10^{-2},10^{-3}\}$ requires the probe to exceed the operating threshold and rank the correct identity first. For target FAR $\phi$ and sorted unknown scores $u_{(1)}\ge u_{(2)}\ge\dots$, the threshold is determined by linear interpolation between bracketing unknown scores. We also report OSCR-AUC, open-set AUROC, EER, and 95\% identity-clustered bootstrap confidence intervals (2000 resamples).

\section{Results}\label{sec:results}

\subsection{Closed-Set Accuracy Is Saturated}

Both embeddings are essentially perfect at the closed-set task. Rank-1 is $99.53\pm0.42\%$ for the hybrid and $99.86\pm0.13\%$ for the foundation model, with verification EER of $0.18\%$ and $0.07\%$.

\subsection{Open-Set Identification}

\begin{table}[h]
\centering
\caption{Open-set identification, plain cosine, oracle threshold.\label{tab:headline}}
\begin{tabularx}{0.8\textwidth}{Xcc}
\toprule
\textbf{Metric} & \textbf{Hybrid CNN--ViT} & \textbf{MegaDescriptor-L}\\
\midrule
Rank-1 (\%) & $99.53\;[99.03,99.86]$ & $99.86\;[99.72,99.97]$\\
Verification EER (\%) & $0.18$ & $0.07$\\
Open-set EER (\%) & $2.77\;[1.90,3.50]$ & $1.63\;[1.02,2.06]$\\
DIR@$10^{-1}$ (\%) & $98.31\;[97.46,99.01]$ & $99.29\;[98.81,99.61]$\\
DIR@$10^{-2}$ (\%) & $96.37\;[95.07,97.64]$ & $98.10\;[97.26,98.92]$\\
DIR@$10^{-3}$ (\%) & $93.59\;[92.18,95.79]$ & $96.07\;[95.32,98.10]$\\
OSCR-AUC & $0.9888\;[0.9827,0.9934]$ & $0.9953\;[0.9926,0.9972]$ \\
Open-set AUROC & $0.9918\;[0.9879,0.9950]$ & $0.9963\;[0.9943,0.9980]$\\
\bottomrule
\end{tabularx}
\end{table}

With plain cosine scoring, the hybrid embedding reaches DIR@FAR of 98.3/96.4/93.6\% at $10^{-1}/10^{-2}/10^{-3}$, and the foundation embedding 99.3/98.1/96.1\%. The important point is that both stay above 90\% at every operating point, including the strict $10^{-3}$ budget. Kumar et al.~\cite{kumar2025} argue that muzzle must be fused with the face, but our results show muzzle images alone support high open-set identification performance on this dataset.

\begin{figure}[H]
\centering
\includegraphics[width=0.45\textwidth]{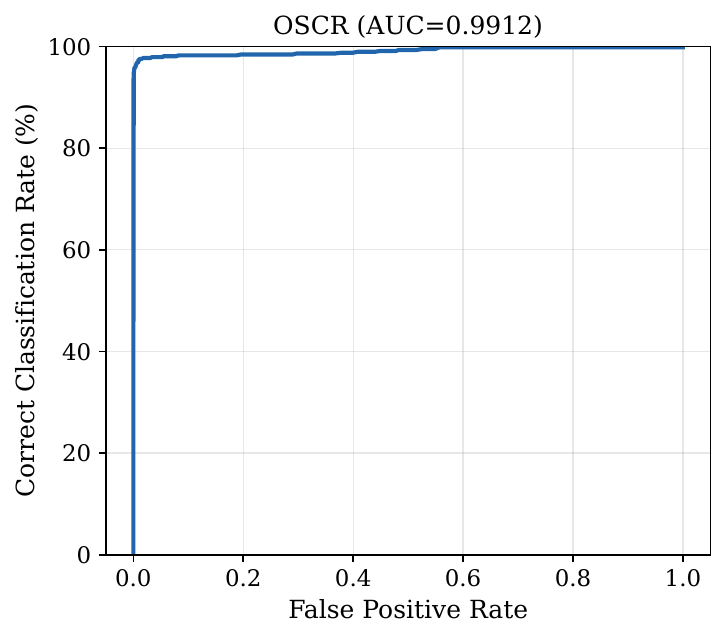}
\caption{Open-set classification rate (OSCR) curve for hybrid CNN-ViT model: plot of true positive rate vs. false positive rate at multiple operating thresholds.}
\label{fig:oscr_hybrid}
\end{figure}

\begin{figure}[H]
\centering
\includegraphics[width=0.45\textwidth]{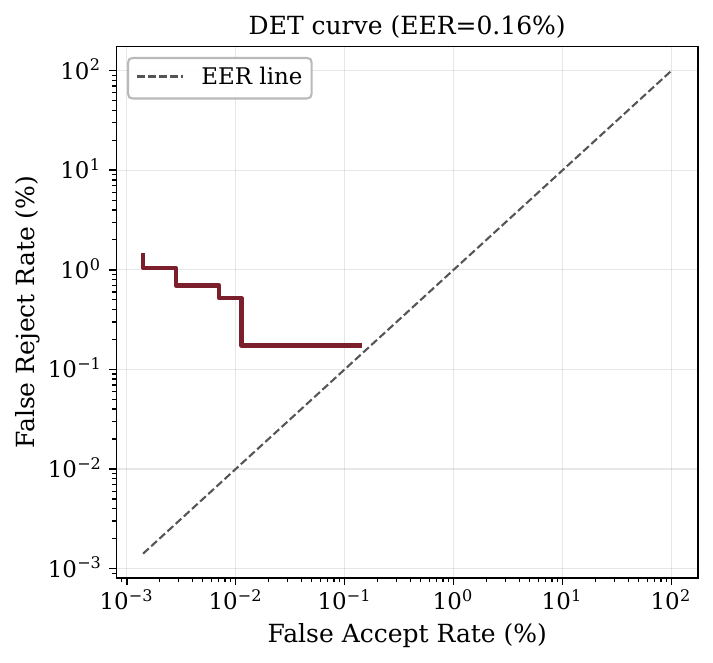}
\caption{Detection-and-identification rate (DIR) curve for hybrid CNN-ViT model at target false-accept rates of $10^{-1}$, $10^{-2}$, and $10^{-3}$.}
\label{fig:det_hybrid}
\end{figure}

\subsection{Evaluation Across Embedding Configurations}

We replace only the embedding with MegaDescriptor-L, and keep enrollment, scoring, and protocol identical; this reproduces the same open-set picture with uniformly stronger numbers, consistent with its large-scale wildlife pretraining. The result we do draw is that the protocol and its qualitative conclusions hold when the embedding configuration changes, so it can absorb future representation-learning advances without being re-derived.

\begin{figure}[H]
\centering
\includegraphics[width=0.55\textwidth]{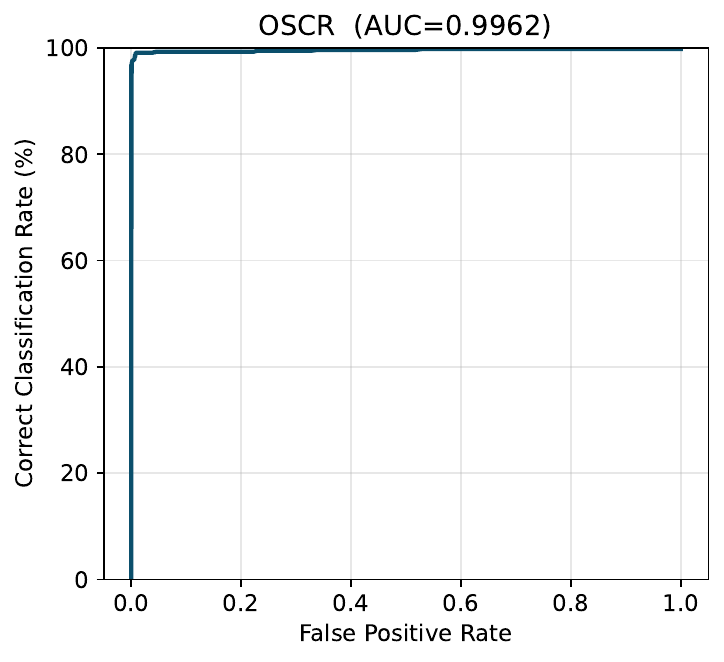}
\caption{Open-set classification rate (OSCR) curve for the MegaDescriptor-L foundation model: demonstrates consistent protocol validity across different embedding architectures.}
\label{fig:oscr_foundation}
\end{figure}

\begin{figure}[H]
\centering
\includegraphics[width=0.55\textwidth]{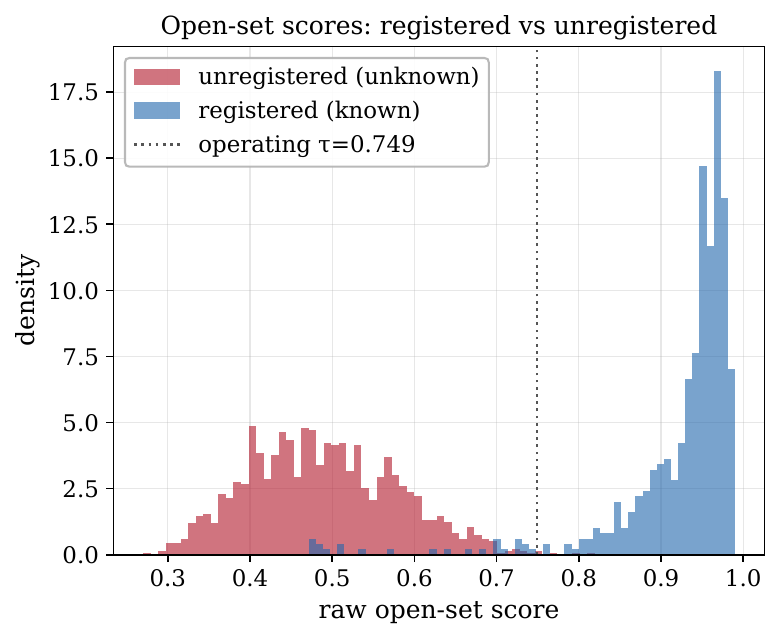}
\caption{Score distribution comparison: hybrid CNN-ViT model with known and unknown sample scores, illustrating the separation between enrolled and unenrolled animals.}
\label{fig:score_dist_hybrid}
\end{figure}

\begin{figure}[H]
\centering
\includegraphics[width=0.55\textwidth]{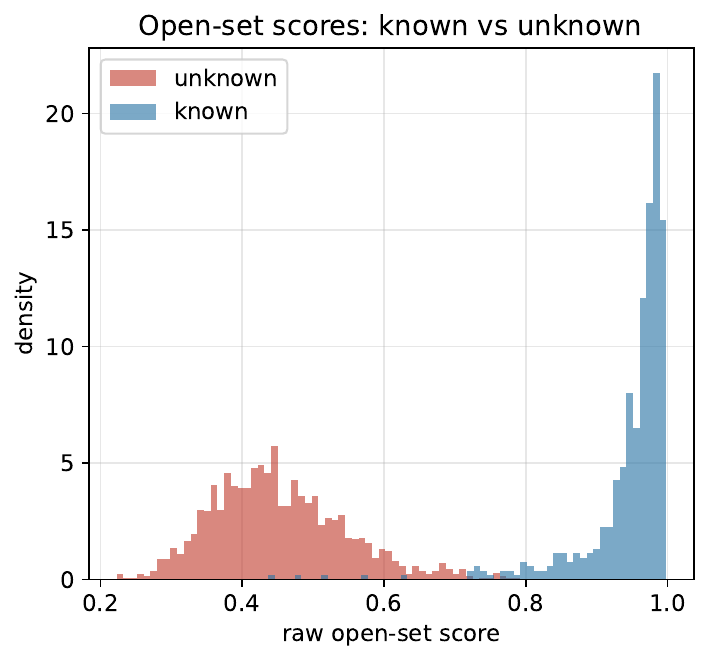}
\caption{Score distribution comparison: MegaDescriptor-L foundation model showing clean separation between known and unknown score distributions.}
\label{fig:score_dist_foundation}
\end{figure}

\subsection{Scoring: Plain Cosine Is Enough}

\begin{table}[h]
\centering
\caption{Scoring-function ablation (foundation embedding; DIR@FAR \%, five-fold mean).\label{tab:scoring}}
\begin{tabularx}{0.7\textwidth}{lcccc}
\toprule
\textbf{Score} & \textbf{$10^{-1}$} & \textbf{$10^{-2}$} & \textbf{$10^{-3}$} & \textbf{OSCR-AUC}\\
\midrule
Plain cosine & 99.29 & \textbf{98.10} & \textbf{96.07} & 0.9953\\
Entropy & 99.29 & 98.10 & 96.04 & 0.9953\\
AS-norm & 99.29 & 97.85 & 94.92 & 0.9953\\
Combined & 99.21 & 97.31 & 93.69 & 0.9955\\
Margin & 99.21 & 97.13 & 92.97 & 0.9954\\
OpenMax & 91.96 & 90.32 & 89.23 & 0.9354\\
\bottomrule
\end{tabularx}
\end{table}

Plain cosine is the best or numerically closest at every operating point, while OpenMax trails by nearly eight points at $10^{-2}$. A large feature-norm gap between known and unknown animals (mean 22.7 versus 19.6, Cohen's $d=2.2$) is present alongside plain cosine's strong performance.

\subsection{Deployment and Consistency}\label{sec:deploy}

Table~\ref{tab:headline} uses an oracle threshold; here we fix the threshold on val-unknown for a $10^{-2}$ target and transfer it unchanged to test-unknown. The hybrid embedding reaches $96.5\%$ detection at a realised FAR of $1.03\%$, close to its target. MegaDescriptor-L reaches $98.2\%$ detection but at a realised FAR of $2.44\%$ -- roughly $2.4\times$ its 1\% target. This is despite MegaDescriptor-L having the better oracle numbers: a stronger score distribution does not guarantee a more transferable threshold. Practically, this argues for recalibrating on farm-specific held-out unknowns rather than transporting a published operating point.

\begin{figure}[H]
\centering
\includegraphics[width=0.65\textwidth]{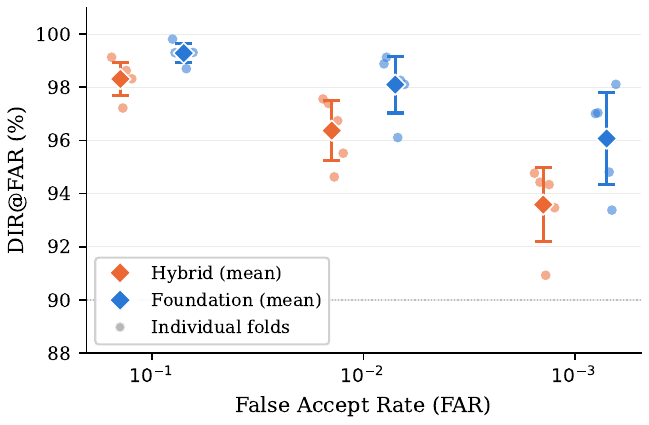}
\caption{Threshold transferability across folds: per-fold performance when threshold calibrated on val-unknown is transferred unchanged to test-unknown, showing deployment consistency.}
\label{fig:per_fold_strip}
\end{figure}

\subsection{Incremental Enrollment}\label{sec:incremental}

The preceding sections evaluate enrolled identities seen during training. In practice, new animals arrive and must be enrolled without retraining. We test this directly: the test-unknown identities (never trained on) are enrolled by averaging $n\in\{1,3,5,8\}$ of their images into a prototype and appending it to the existing gallery while the remaining images serve as queries.

\begin{table}[h]
\centering
\caption{Incremental enrollment of previously unseen identities (no retraining). $n$: enrollment images per new identity.\label{tab:incremental}}
\begin{tabularx}{0.8\textwidth}{cccccc}
\toprule
& $n$ & \textbf{Rank-1 new} & \textbf{DIR@$\tau$ new} & \textbf{Rank-1 orig.} & \textbf{Reject unk.} \\
\midrule
\multirow{4}{*}{Hybrid}
 & 1 & 91.08 & 75.63 & 99.39 & 98.83 \\
 & 3 & 95.94 & 86.22 & 99.39 & 98.76 \\
 & 5 & 96.35 & 88.28 & 99.38 & 98.80 \\
 & 8 & 97.07 & 89.37 & 99.38 & 98.84 \\
\midrule
\multirow{4}{*}{Foundation}
 & 1 & 92.61 & 77.54 & 99.80 & 98.86 \\
 & 3 & 96.46 & 86.38 & 99.77 & 98.86 \\
 & 5 & 96.93 & 88.37 & 99.78 & 98.86 \\
 & 8 & 97.27 & 88.40 & 99.80 & 98.86 \\
\bottomrule
\end{tabularx}
\end{table}

A single enrollment image already yields Rank-1 above 91\% for both embeddings, though detection at the operating threshold is markedly lower at that point. By eight images, Rank-1 reaches 97.1\%/97.3\%. Crucially, the original gallery is unaffected; Rank-1 on the original probes stays above 99.3\%, and the rejection rate on remaining unknowns stays above 98.7\%.

\begin{figure}[H]
\centering
\includegraphics[width=0.55\textwidth]{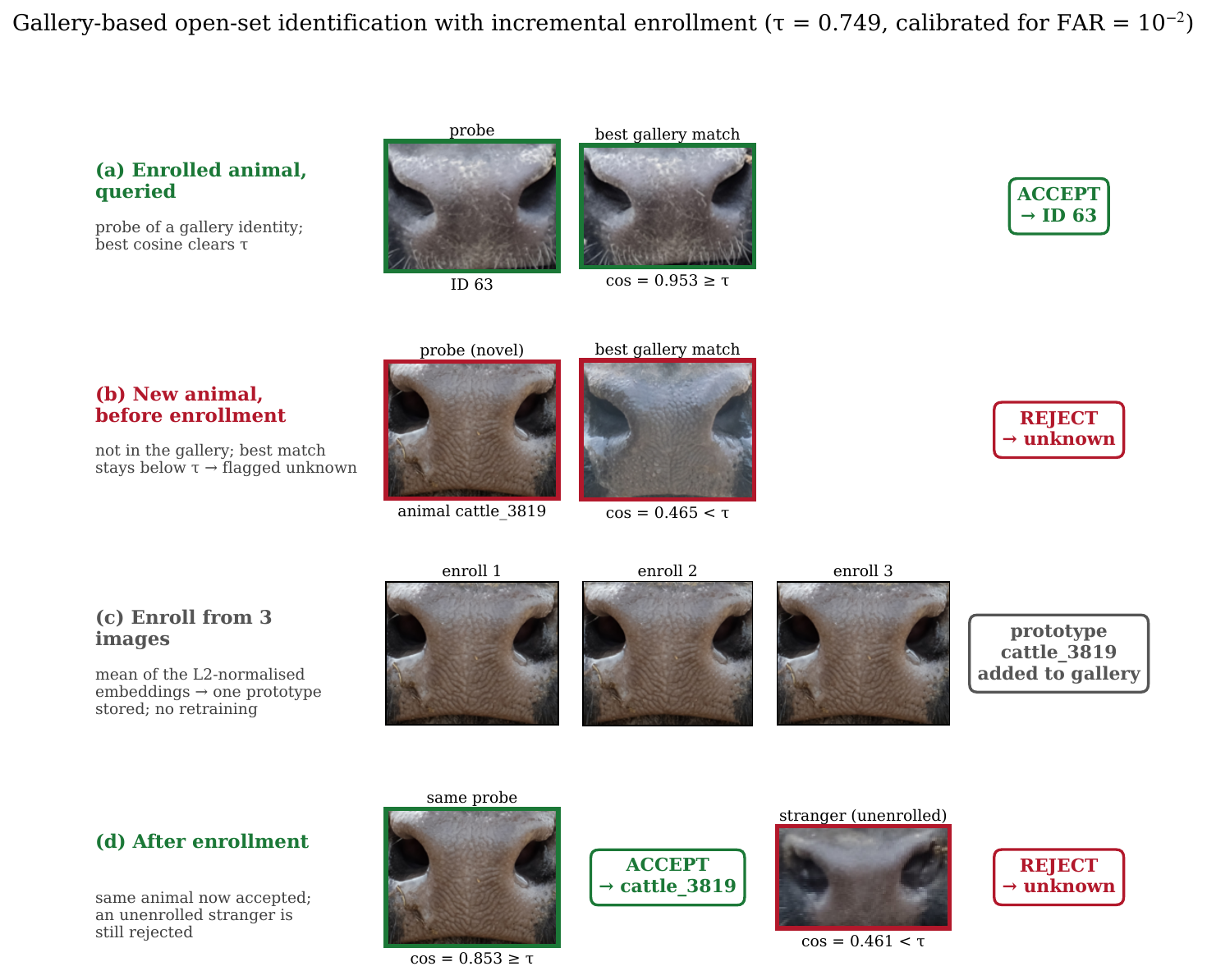}
\caption{Incremental enrollment demonstration: gallery-based enrollment with no retraining enables adding new identities while maintaining performance on the existing gallery.}
\label{fig:enrollment_demo}
\end{figure}

\section{Discussion}\label{sec:discussion}

Framing muzzle identification as open-set recognition, and testing it under a leakage-controlled protocol, gives a system that both identifies enrolled animals and rejects strangers at strict false-accept rates, for two embedding configurations that differ in both backbone and loss. Kumar et al.~\cite{kumar2025} argue the muzzle must be fused with the face, but our results show muzzle images alone supported high open-set performance on this single-herd dataset, not a general claim against fusion. The contribution is the formulation and the protocol, and the embeddings demonstrate rather than define them.

The threshold-transfer result (Section~\ref{sec:deploy}) is the most practically consequential finding: MegaDescriptor-L had the stronger oracle numbers and the worse calibration. This argues for reporting realised FAR under transferred thresholds as a primary result in open-set livestock biometrics, and for treating on-farm recalibration as expected rather than optional. On the scoring side, plain cosine performing comparably to more elaborate alternatives has a practical deployment reading: cosine scoring against $M$ prototypes is one $M\times512$ matrix--vector product, scaling linearly and remaining feasible into the tens of thousands of identities before approximate search is warranted.

The incremental-enrollment and gallery-growth experiments connect the open-set formulation to deployment, though they establish less than a full incremental-enrollment claim on their own. Test-unknown identities never seen during training reach high Rank-1 from a small number of images, the existing gallery is essentially unaffected by expansion, and realised FAR under a fixed threshold drifts upward as the gallery grows.

We do not claim architectural novelty; its parts are standard. We did not run component-level ablations isolating the fusion, the gate, and the loss. Limitations on the present result: single herd, camera, and acquisition period, with breed diversity limited to Angus and two crosses; identity-disjoint splitting removes identity leakage but not possible acquisition-session, background, or camera-signature cues, which we did not separately audit; FAR=$10^{-3}$ resolution-limited to roughly 1--2 expected false accepts per partition; no isolated component ablations; gallery growth evaluated to 179 identities under one enrollment ordering, not thousands or multiple orderings; and no evaluation of age-related pattern stability, muzzle soiling or moisture, presentation/spoofing attacks, or prospective field deployment.

\begin{figure}[H]
\centering
\includegraphics[width=0.65\textwidth]{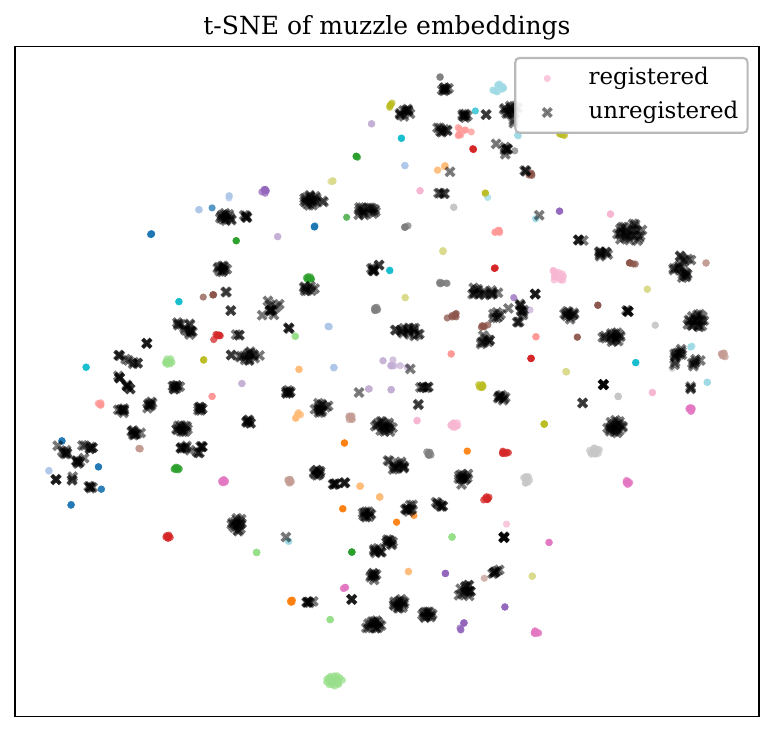}
\caption{t-SNE visualization of embedding space: hybrid CNN-ViT model showing known identities (colored clusters) and unknown samples (gray), demonstrating the separation necessary for open-set rejection.}
\label{fig:tsne_embedding}
\end{figure}

\begin{figure}[H]
\centering
\includegraphics[width=\textwidth]{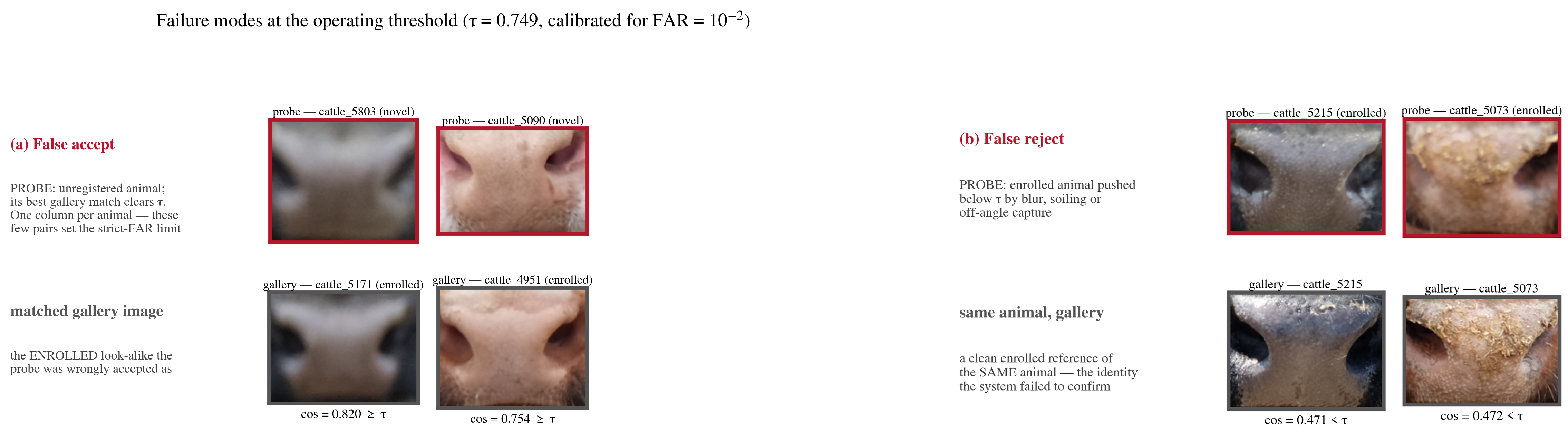}
\caption{Failure modes analysis: representative examples of misidentification and false rejections from the hybrid model, highlighting challenging cases with poor image quality, occlusion, or extreme poses.}
\label{fig:failure_modes}
\end{figure}

\section{Conclusions}

We reformulated cattle muzzle identification as an open-set, gallery-based problem that rejects unenrolled animals and enrolls new ones without retraining, and introduced a leakage-controlled evaluation protocol. Instantiated with a hybrid CNN--ViT embedding and, independently, the MegaDescriptor-L foundation model, the framework identifies enrolled animals at 94--98\% detection across false-accept budgets from $10^{-1}$ to $10^{-3}$ under oracle thresholds, and at 96.5\%/98.2\% detection with realised FAR of 1.03\%/2.44\% under thresholds calibrated on held-out unknowns and transferred unchanged. Gallery-based enrollment adds previously unseen identities from as few as one to eight images without retraining or degrading the existing gallery. Within this single-herd dataset, muzzle images alone supported open-set identification without face fusion, and plain cosine scoring gave the strongest or comparable performance among the methods evaluated. We have released the protocol so that future muzzle systems can be judged on their ability to abstain, not only to rank.

\section{Acknowledgments}
The authors acknowledge the use of GPU computational resources provided through RunPod and Google Colab, which were used to conduct the experiments in this study.

\section{Author Contributions}

Conceptualization, methodology, original draft preparation, and validation: Lalit BC. Methodology, data curation, data analysis, and validation: Dharmendra Singh Chaudhary. Validation, editing, and draft preparation: Shovit Nepal. All authors have read and agreed to the published version of the manuscript.

\section{Funding}

This research received no external funding.

\section{Data Availability}

The original Beef Cattle Muzzle/Nose-print Database is public and available at \url{https://zenodo.org/records/6324361}~\cite{xiong2022}. The evaluation protocol, split files, duplicate-identity list, and code are available at \url{https://github.com/Dharmendra016/Muzzlenet} under the MIT License.

\section{Conflicts of Interest}

The authors declare no conflicts of interest. 

\bibliographystyle{plain}

\end{document}